\documentclass[letterpaper, 10 pt, conference]{ieeeconf}  
\IEEEoverridecommandlockouts                              

\title{Safety Assurances for Human-Robot Interaction via \\Confidence-aware Game-theoretic Human Models}

\author{Ran Tian$^*$, Liting Sun$^*$, Andrea Bajcsy$^*$, Masayoshi Tomizuka, and Anca D. Dragan
\thanks{$^{*}$Equal contribution. All authors are with UC Berkeley. {\tt\footnotesize \{rantian, litingsun, abajcsy, tomizuka, anca\}@berkeley.edu}}
}

\usepackage{tabularx}
\usepackage{float}
\usepackage{epsfig}
\usepackage{epstopdf}
\usepackage{bbm}
\usepackage{mathtools,xparse}

\usepackage{epsfig} 
\usepackage{amsmath} 
\usepackage{amssymb}  
\usepackage{bm}
\usepackage{color}
\usepackage{etoolbox}

\usepackage{comment}
\usepackage{diagbox}
\usepackage[ruled, lined, longend, linesnumbered]{algorithm2e}
\usepackage{booktabs,multirow}
\usepackage{bigstrut}
\usepackage{cite}
\usepackage{subcaption,booktabs}
\usepackage{color, colortbl}

\usepackage{lipsum}

\usepackage{hyperref}
\usepackage{balance}

\usepackage{amsfonts}
\usepackage{mathtools}
\usepackage{xcolor}

\usepackage{graphicx}
\usepackage[capitalise]{cleveref}
\usepackage{hyperref}

\usepackage{amsmath,amssymb,stmaryrd,mathtools}

\definecolor{orange(sae/ece)}{rgb}{1.0, 0.49, 0.0}

\newcommand{\xH}{x_\mathrm{H}}

\newcommand{\uH}{u_\mathrm{H}}

\newcommand{\thor}{T}

\newcommand{\jointstate}{x}
\newcommand{\jointdyn}{f}

\newcommand{\xrel}{x_\mathrm{rel}}
\newcommand{\reldyn}{f_\mathrm{rel}}
\newcommand{\rel}{\mathrm{rel}}

\newcommand{\xR}{x_\mathrm{R}}

\newcommand{\uR}{u_\mathrm{R}}
\newcommand{\uRopt}{u^*_\mathrm{R}}

\newcommand{\uRtraj}{\mathbf{\uR}}
\newcommand{\uHtraj}{\mathbf{\uH}}
\newcommand{\obsuRtraj}{\mathbf{\hat{u}_\mathrm{R}}}
\newcommand{\obsuHtraj}{\mathbf{\hat{u}_\mathrm{H}}}

\newcommand{\uRtrajopt}{\mathbf{\uRopt}}

\newcommand{\tildeuHtraj}{\mathbf{\tilde{u}_\mathrm{H}}}

\newcommand{\uHtrajset}{\mathbb{U}_\mathrm{H}}

\newcommand{\uHsetapprox}{\widetilde{\mathcal{U}}_\mathrm{H}}
\newcommand{\targetfun}{\ell}

\newcommand{\vfun}{V}
\newcommand{\brs}{\mathcal{V}} 
\newcommand{\approxbrs}{\widetilde{\mathcal{V}}} 
\newcommand{\truebrs}{\mathcal{V}^*} 

\definecolor{orange(sae/ece)}{rgb}{1.0, 0.49, 0.0}
\definecolor{light_gray}{rgb}{0.9, 0.9, 0.9}
\definecolor{medium_gray}{rgb}{0.6, 0.6, 0.6}

\begin{document}
\maketitle

\begin{abstract}
An outstanding challenge with safety methods for human-robot interaction is reducing their conservatism while maintaining robustness to variations in human behavior. 
In this work, we propose that robots use confidence-aware game-theoretic models of human behavior when assessing the safety of a human-robot interaction. 
By treating the influence between the human and robot as well as the human's rationality as unobserved latent states, we succinctly infer the degree to which a human is following the game-theoretic interaction model. We leverage this model to restrict the set of feasible human controls during safety verification, enabling the robot to confidently modulate the conservatism of its safety monitor online. Evaluations in simulated human-robot scenarios and ablation studies demonstrate that imbuing safety monitors with confidence-aware game-theoretic models enables both safe and efficient human-robot interaction. Moreover, evaluations with real traffic data show that our safety monitor is less conservative than traditional safety methods in real human driving scenarios.
\end{abstract}



%
\IEEEpeerreviewmaketitle

\section{Introduction}

\begin{figure}
    \centering
    \includegraphics[width=\columnwidth]{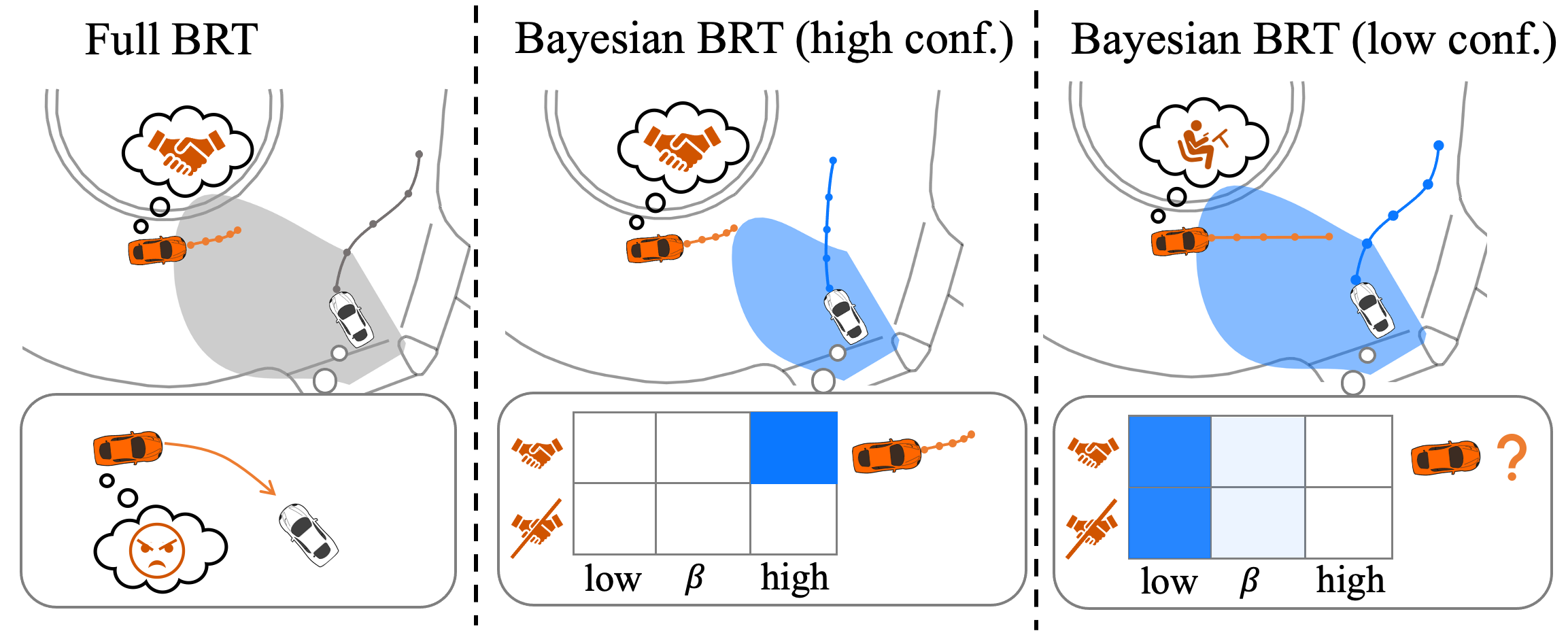}
    \caption{Robot car (white) merges into a round-about with a nearby human-driven car (orange). (left) Human accommodates for robot, but robot is overly conservative and protects against the full backwards reachable tube (BRT). (center) Our Bayesian BRT infers how the human is influenced by the robot and shrinks the set of unsafe states. (right) When the human does not behave according to the model, the robot detects this and automatically reverts to the full BRT.}
    \label{fig:front_fig}
    \vspace{-2em}
\end{figure}

We focus on maintaining safety in highly dynamic human-robot interactions, such as when an autonomous car merges into a roundabout with an oncoming human-driven vehicle (Fig. \ref{fig:front_fig}). While planning approaches incorporate safety constraints in diverse ways \cite{schwarting2018planning}, \emph{safety monitors} have emerged as a desirable additional layer of safety. These methods allow the planner to guide the robot, but compute when imminent collisions would happen and take over control to steer the robot away from danger.

Crucial to these safety monitors is a method for detecting imminent collisions. Typically, this is based on \emph{worst-case} reasoning. A predominant approach, \emph{backwards reachability analysis} \cite{mitchell2005time, leung2020infusing}, treats the human-robot interaction as a zero-sum collision-avoidance game, protecting the robot against \emph{any} controls the human might execute. This leads to safety monitors that do maintain safety, but inhibit the robot's ability to make progress by intervening excessively.

We thus seek a way of making safety monitors less conservative, while still being effective at their primary job---maintaining safety. What makes this challenging is that the moment the zero-sum game assumption is replaced with any human behavior model, the model might be wrong, leading to loss of safety. Our idea is to mediate this issue in two ways: 1) still use a zero-sum collision avoidance game, but instead of allowing the human \emph{any} controls, we use a human behavior \emph{model} to \emph{restrict} set of controls we safeguard against, eliminating those that the model deems very improbable; and 2) detect online \emph{how well the model fits the human}, and use this to adapt the restriction; at the extreme, when the model is completely wrong, our monitor should go back to protecting against any human controls.

Two questions still remain: what human model to use, and how to detect when it is wrong. While models that treat the human as acting in isolation and ignoring the robot are popular \cite{ziebart2009planning, karasev2016intent, rudenko2017predictive}, they are still very conservative: if the planner tries to merge in front of the human, the safety monitor based on these ``human-in-isolation'' models would intervene to prevent it, because it has no confidence in the human reacting to the robot and making space---also known as the ``frozen robot'' problem \cite{trautman2010unfreezing}. For this reason, prior work in planning has introduced models based on \emph{general-sum} games between the human and robot, which account for the human's influence on the robot, but also for the robot's influence on the human \cite{sadigh2016planning, schwarting2019social, fisac2019hierarchical,tian2021}. We propose to use such models in safety monitoring too, as a way to restrict the set of controls that the robot safeguards against. In our method, the robot performs backwards reachability analysis but does not worry about human controls that are outside of the bounds of what the general sum game deems likely.

While this reduces the conservatism, no model is perfect and relying solely on the human model might remove controls that the human actually ends up executing. To detect if the model's predictive performance is degrading and increase conservatism when it does, our approach uses online observations of human behavior to assess the quality of the model and adapt the control bound restriction. Building on prior work in safe planning \cite{fisac2018probabilistically}, we achieve automatic adaptation by treating human behavior as observations of the human's \emph{rationality} level in the general-sum game. When the human starts executing less-probable controls, the model treats this behavior as more noisy, and deems more future human controls to be likely to occur than before. 
In turn, this propagates to a larger human control bound in our backwards reachability analysis and a more conservative safety monitor.

We test our approach in simulated interactions, as well as on real human driving data. Our results suggest that it can effectively enable robots to modulate the conservatism of their safety monitors, ultimately leading to more efficient (i.e., higher reward) behaviors that still maintain safety.

\section{Related Work}

\noindent\textbf{Safety for robots operating around humans.}
Forward reachability methods have been used to compute the set of states where other agents could be in the future \cite{althoff2011set, liu2017provably}, after which the robot plans to avoid this set. While safe, these methods often lead to overly-conservative robot behaviors especially in close-proximity interactive scenarios. Prior work used empirical \cite{driggs2018robust} and ``human-in-isolation'' models \cite{bajcsy2020robust} to obtain restrictions on human controls and, ultimately, their forward reachable set. However, these methods do not inherently account for the robot's ability to take safety maneuvers in response to the human behavior, introducing additional conservatism.
In contrast, backwards reachability methods are grounded in zero-sum dynamic games \cite{mitchell2005time, leung2020infusing} which encode the robot's ability to enact safety controls. 
While generally less conservative than full forward reachability, this approach still often suffers from falsely flagging safe states as unsafe because of the full control authority and adversarial nature of the human model. Recent work has attempted to reduce this conservatism by restricting the set of human controls during safety analysis through data-driven human trajectory forecasts \cite{li2020prediction}. However, this approach blindly trusts the data-driven forecasts and cannot detect model errors; when the quality of the data-driven forecasts degrade, so do the predicted set of human controls, ultimately compromising the safety monitor. 


\smallskip
\noindent \textbf{Structured human decision-making models.}
``Human-in-isolation'' models---whereby the human is treated as behaving independently of how other agents nearby behave--- have been widely studied and applied in the navigation domain \cite{ziebart2009planning, karasev2016intent, rudenko2017predictive}. Recent work has developed model confidence monitors for this class of models \cite{fisac2018probabilistically, bobu2020quantifying}, enabling robots to detect if and when their human models are misspecified.
Unlike human-in-isolation models, theory of mind models capture how humans account for the behavior of others when choosing their actions \cite{baker2014modeling, carruthers1996theories, gergely2003teleological, sodian2004infants}. Recent work has shown the effectiveness of using such models---specifically, general-sum Stackelberg games \cite{von2010market}---in the context of autonomous driving
\cite{sadigh2016planning, schwarting2019social, fisac2019hierarchical}.
In our work, we leverage such game-theoretic human models and introduce a confidence-aware monitor for this modelling paradigm. 



\section{Background}
\noindent \textbf{Hamilton-Jacobi reachability.}
Our method is based on Hamilton-Jacobi (HJ) reachability analysis \cite{mitchell2005time, margellos2011hamilton}, a mathematical formalism for quantifying the performance and safety of multi-agent dynamical systems. It has been successfully utilized in a range of safety-critical applications such as multi-vehicle planning \cite{chen2015safe, dhinakaran2017hybrid}, multi-player reach-avoid games \cite{huang2011differential, margellos2011hamilton}, and autonomous driving under occlusions \cite{zhang2021safe} due to its ability to handle general nonlinear dynamics, flexibility to represent unsafe sets of arbitrary shapes, and ability to synthesize safety-preserving robot controllers. 

In this work, we use reachability analysis to compute a \textit{backward reachable tube} (BRT), $\brs(\tau)$, given an unsafe set of states $\mathcal{L}$ (e.g., all states where the human and robot are in collision). 
Intuitively, $\brs(\tau)$ is the set of states from which if system trajectories start, they are guaranteed to enter into the unsafe set of states within a time horizon of $\tau$ despite the robot's best effort to avoid the unsafe set. 

Let the dynamics of the human-robot system evolve via $\dot{\jointstate} = \jointdyn(\jointstate, \uH, \uR)$ where $\jointdyn$ is assumed to be uniformly continuous in time and Lipschitz continuous in $\jointstate$ for fixed $\uH$ and $\uR$.
Here, $\jointstate := [\xH, \xR]^\top \in \mathcal{X}$ is the joint state\footnote{Other state-space representations can be used. In Section~\ref{sec:experimental_setup} we use the relative state between the human and robot to reduce state dimensionality.} of the human and robot, and $\uH \in \mathcal{U}_\mathrm{H}$ and $\uR \in \mathcal{U}_\mathrm{R}$ are the human's and robot's inputs, respectively. We also assume that the state of both agents can be accurately sensed at all times. 

To ensure robustness to the possible---including worst-case---behaviors of the human agent, the computation of the BRT is formulated as a zero-sum differential game between the robot and human. 
The optimal value of this game can be obtained by solving the final-value Hamilton-Jacobi-Isaacs Variational-Inequality (HJI-VI) via dynamic programming: 
    \begin{align}
        \min\{ &\frac{\partial \vfun(\jointstate, \tau)}{\partial \tau} +  H(\jointstate, \tau, \nabla \vfun(\jointstate, \tau)), \targetfun(\jointstate) - \vfun(\jointstate, \tau)\} = 0, \nonumber\\
        &\vfun(\jointstate, 0) = \targetfun(\jointstate), ~\tau \in [-\thor, 0] ,
        \label{eq:hji_vi}
    \end{align}
where $\nabla \vfun(\jointstate, \tau)$ is the spatial derivative of the value function and  $\targetfun(\jointstate)$ is the implicit surface function encoding the set of unsafe states: $\mathcal{L} = \{ \jointstate : \targetfun(\jointstate) \leq 0\}$.
The Hamiltonian $H$ encodes the effect of the dynamics, robot, and human control on the resulting value and is defined as:
 \begin{equation}
     H(\jointstate, \tau, \nabla \vfun(\jointstate, \tau)) = \max_{\uR \in \mathcal{U}_\mathrm{R}} \min_{\uH \in \mathcal{U}_\mathrm{H}} \nabla \vfun(\jointstate, \tau)^\top \jointdyn(\jointstate, \uH, \uR).
        \label{eq:hamiltonian}
\end{equation}


After computing the value function $\vfun(\jointstate, \tau)$ backwards in time over $\tau \in [-\thor, 0]$, we can obtain the BRT at any time $\tau$ by looking at the sub-zero level set of the value function:
    \begin{equation}
        \brs(\tau) := \{\jointstate : \vfun(\jointstate, \tau) \leq 0\}.
        \label{eq:brs}
    \end{equation}
This encodes the set of initial (joint) states from which there does not exist a dynamically-feasible safety control for the robot to perform to avoid the human. 
HJ reachability also synthesizes the robot's optimal safety-preserving control:
\begin{equation}
    \uR^*(\jointstate, \tau) = \arg\max_{\uR \in \mathcal{U}_\mathrm{R}} \min_{\uH \in \mathcal{U}_\mathrm{H}} \nabla \vfun(\jointstate, \tau)^\top \jointdyn(\jointstate, \uH, \uR),
    \label{eq:hj_opt_control}
\end{equation} 
which can be used in a least-restrictive fashion by only being applied at the boundary of the unsafe set \cite{bansal2017hamilton}.

Note that in \eqref{eq:hamiltonian}, the agents are traditionally modelled as optimizing with respect to \textit{all} of their dynamically-feasible control inputs, $\mathcal{U}_\mathrm{i}, i \in \{\mathrm{H}, \mathrm{R}\}$, resulting in an overly conservative BRT. In this work, we aim to reduce the conservatism (i.e., size of $\brs(\tau)$) of this safety monitor by detecting emergent leader-follower roles in human-robot interaction and restricting the human's controls $\mathcal{U}_\mathrm{H}$ accordingly. 

\smallskip
\noindent \textbf{General-sum Stackelberg games.}
We model humans as acting according to a discrete-time, general-sum Stackelberg game \cite{von2010market} with the robot wherein each agent takes on the role of being either a ``leader'' or a ``follower''. A ``leader'' maximizes their reward over time subject to the ``follower'' who must plan their trajectory in response. Intuitively, the leader aims to influence the follower and the follower tends to accommodate the leader. This modelling paradigm is well-suited to capture dynamic interactions such as merging or lane-changing and, unlike zero-sum games, can encode unique high-level objectives for each agent. 

Let the human's discrete-time control trajectory over the time horizon $\thor$ be denoted by $\uHtraj = [\uH^0, \hdots, \uH^\thor]^\top$ and the robot's discrete-time control trajectory be $\uRtraj = [\uR^0, \hdots, \uR^\thor]^\top$. 
Instead of assuming the human is a perfectly rational player, we model them a noisily-optimal player; this is well-suited for human models obtained via inverse reinforcement learning \cite{ziebart2008maximum, waugh2010inverse, levine2012continuous} and naturally accounts for model inaccuracies and noisy human behavior.
In an open-loop Stackelberg game, a noisily-optimal \textit{follower} human chooses their control trajectory from a distribution conditioned on the leader robot's control trajectory:
    \begin{equation}
    \begin{aligned}
        P(\uHtraj \mid \jointstate^0, \uRtraj; \beta) = \frac{1}{Z_1} e^{\beta R_{\mathrm{H}}(\jointstate^0, \uHtraj, \uRtraj)},  \\
    \end{aligned}
    \label{eq:olgame_hfollower_rlead}
    \end{equation}
where $Z_1 := \int_{\tildeuHtraj } e^{\beta R_{\mathrm{H}}(\jointstate^0, \tildeuHtraj, \uRtraj)} d\tildeuHtraj$ is the partition function when the human is a follower and the human's cumulative reward is $R_\mathrm{H}(\jointstate^0, \uHtraj, \uRtraj) := \sum^\thor_{\tau = 0} r_\mathrm{H}(\jointstate^\tau, \uH^\tau, \uR^\tau)$ where $r_\mathrm{H}(\cdot, \cdot, \cdot)$ is the instantaneous reward. 
    
However, in reality, the human could assume the role of either a leader or follower during an interaction with the robot.
Specifically, a human who assumes the role of a \textit{leader} will draw their control trajectory from the distribution:
    \begin{equation}
    \begin{aligned}
       P(\uHtraj \mid \jointstate^0; \beta) = \frac{1}{Z_2} e^{\beta R_{\mathrm{H}}(\jointstate^0, \uHtraj, \uRtrajopt(x^0, \uHtraj))},
    \end{aligned}
    \label{eq:olgame_hlead_rfollow}
    \end{equation}
where $Z_2 := \int_{\tildeuHtraj } e^{\beta R_{\mathrm{H}}(\jointstate^0, \tildeuHtraj, \uRtrajopt(x^0, \tildeuHtraj))} d\tildeuHtraj$ is the partition function and $\uRtrajopt(x^0, \uHtraj)$ is the robot follower's best response to the human's control trajectory $\uHtraj$ and is defined as: 
    \begin{equation}
    \begin{aligned}
       \uRtrajopt(\jointstate^0, \uHtraj) = \max_{\uRtraj} \quad R_{\mathrm{R}}(\jointstate^0, \uRtraj, \uHtraj),\\
       \textrm{s.t.} \quad \jointstate^{\tau+1} = \tilde{f}(\jointstate^{\tau}, \uR^\tau, \uH^\tau), \quad \tau \in \{0, \hdots, \thor\},
    \end{aligned}
    \label{eq:olgame_rfollow}
    \end{equation}
where the robot's running reward is denoted by $R_{\mathrm{R}}(\jointstate^0, \uRtraj, \uHtraj) := \sum^\thor_{\tau = 0} r_\mathrm{R}(\jointstate^\tau, \uR^\tau, \uH^\tau)$. 

Finally, the parameter $\beta \in [0, \infty)$ encodes the human's rationality and governs how optimally the human behaves according to their objective; as $\beta \rightarrow 0$, the human appears ``irrational'', choosing their trajectory uniformly at random and ignoring any modeled structure, while $\beta \rightarrow \infty$ models the human as a perfect optimizer of the game. 


\section{Confidence-aware Role Inference for Safe Human-Robot Interaction}
We propose that robots estimate the degree to which humans abide by general-sum interaction models and adapt their safety monitors accordingly.
By reducing the unsafe states proportional to the observed influence between the human and robot, the robot can smoothly shift between less conservative safety monitors when the human's behavior is well-explained by the general-sum model, and the full worst-case safety monitor when the human model degrades.

\smallskip 
\noindent \textbf{Role-parameterized human model.}
We treat the role of the human in the general-sum game as well as their apparent rationality as hidden states. 
Let $\lambda \in \{\mathrm{follow}, \mathrm{lead}\}$ be a discrete latent variable which encodes the role of the human as either a follower or leader. 
To assess if the observed human behavior matches our general-sum model, we follow prior work on ``human-in-isolation'' models in reinterpreting the human's rationality parameter $\beta$ as an indicator of model confidence \cite{fisac2018probabilistically, bobu2020quantifying}. However, to infer both the role of the human as well as the degree to which the human is playing a general-sum game at all, the robot jointly infers $(\lambda, \beta)$ given observations. 

Online, after observing the current joint state $\jointstate^0$, the robot solves two open-loop general-sum Stackelberg games where the human swaps roles.
Our final stochastic human model is:
    \begin{equation}
    P(\uHtraj \mid \jointstate^0, \uRtraj; \lambda, \beta) =
    \begin{cases}
        P(\uHtraj \mid \jointstate^0, \uRtraj; \beta), & \lambda = \mathrm{follow} \\
        P(\uHtraj \mid \jointstate^0; \beta), & \lambda = \mathrm{lead}
    \end{cases}    
    \label{eq:conf_aware_human_role_model}
    \end{equation}
where each sub-model is computed via solving \eqref{eq:olgame_hlead_rfollow} and \eqref{eq:olgame_hfollower_rlead}.

\textit{Remark:} Computing the partition function in either of the two models from \eqref{eq:conf_aware_human_role_model} requires integrating over the space of possible trajectories, $\Pi$, which is infinite. Furthermore, due to the nested optimization in the partition function when the human is a leader, approximation techniques based on Laplace approximation are not applicable. Instead, a finite choice set ($\widetilde{\Pi} \subset \Pi$) sampled from a background distribution is often exploited \cite{finn2016guided, wu2020efficient}. In our experiments, we use a simple generative model based on second order polynomials to synthesize the human-driven car's likely acceleration and steering profiles at each time step to construct $\widetilde{\Pi}$. 

\smallskip
\noindent \textbf{Confidence-aware role inference. }
At each time step, the robot maintains a joint belief over the leader-follower roles and model confidence.
Let the time horizon over which we observe trajectory snippets to be $\thor$ seconds and the current time to be denoted by $0$. Starting from the past state $\hat{x}$, the robot observes its executed trajectory $\obsuRtraj := [\uR^{-\thor}, \hdots, \uR^0]$ over the time interval $[-\thor,0]$ and the human's behavior $\obsuHtraj$ over the same interval. Using these and the model \eqref{eq:conf_aware_human_role_model}, the robot updates its belief about $(\lambda, \beta)$ via a Bayesian update: 
\begin{equation}
   b'(\lambda, \beta \mid \hat{x}, \obsuHtraj, \obsuRtraj) \propto P(\obsuHtraj \mid \hat{x}, \obsuRtraj; \lambda, \beta) b(\lambda, \beta).
\end{equation}
Since we will use the belief to modify the set of unsafe states, its critical that this update be performed extremely fast. In theory $\beta \in [0, \infty)$, resulting in an update that would be computationally prohibitive; however, in Section~\ref{sec:sim_human_robot_results} we demonstrate that maintaining a belief over a relatively small set of $\beta$ values performs favorably. 

\smallskip
\noindent \textbf{Online update of the safety monitor.}
To modulate the conservatism of the BRT, we will look to the current belief over the human's role and the model confidence to weight the likelihood of future human control trajectories. 
Specifically, we marginalize over $(\lambda, \beta)$ according to the current belief $b(\lambda, \beta)$ to obtain a distribution over human trajectories starting from the current state $\jointstate^0$:
\begin{equation}
\hspace{-0.2cm}  P(\uHtraj \mid \jointstate^0, \uRtraj) = \mathbb{E}_{\lambda, \beta \sim b(\lambda, \beta)} \big[P(\uHtraj \mid \jointstate^0, \uRtraj; \lambda, \beta)\big].
    \label{eq:htraj_marginal_dist}
\end{equation} 
This model enables us to determine which future control profiles the robot can expect to see from the human in response to the robot's own motion plan. To leverage our modelled structure and reduce conservatism, we prune away human control trajectories that are sufficiently unlikely under the marginal distribution; let this set be  
 $\uHtrajset(\uRtraj) := \{\uHtraj : P(\uHtraj \mid \jointstate^0, \uRtraj) > \epsilon\}$, where $\epsilon$ is a hyperparameter that controls how many unlikely trajectories get pruned. 

Finally, we must transform the set of likely control trajectories into instantaneous control bounds to compute the unsafe set. 
For simplicity, we have the robot safeguard against the maximum and minimum controls\footnote{A more computationally costly time-varying restriction is also possible.} the human could execute during \textit{any} of the likely trajectories in $\uHtrajset(\uRtraj)$. Although this approach introduces some conservatism into the estimated human control bounds (and the resulting unsafe states), it does provide an additional layer of robustness to the exact way in which people execute their local motions. Our results in Section~\ref{sec:sim_human_robot_results} additionally indicate useful reduction in the size of the unsafe sets for a variety of scenarios. 
Let the restricted set of human controls be $\uHsetapprox := \big[\underline{u}_\mathrm{H}(\uRtraj), \overline{u}_\mathrm{H}(\uRtraj)\big]$ where $\underline{u}_\mathrm{H}(\uRtraj) = \min_{\tau \in \{0,\hdots, \thor\}} \uHtrajset(\uRtraj)$ is the lower control bounds and $\overline{u}_\mathrm{H}(\uRtraj) = \max_{\tau \in \{0,\hdots, \thor\}} \uHtrajset(\uRtraj)$ the upper.

The final set of restricted human controls $\uHsetapprox$ is ultimately used in the computation of the BRT when evaluating the Hamiltonian in Eq.~\eqref{eq:hamiltonian}.
By solving the HJI-VI with these restricted human control bounds, obtaining the corresponding value function, and computing the unsafe states via the sub-zero level set from \eqref{eq:brs}, we finally obtain the restricted \textit{Bayesian BRT}, $\approxbrs(\tau)$. Note that this set of unsafe states is recomputed\footnote{In practice, we pre-computed a bank of BRTs using various control bounds and query the BRT associated with $\uHsetapprox$ online.} after each belief update, since the belief affects the human's control bounds used in the reachability computation and influences the size and shape of $\approxbrs(\tau)$. 

Ultimately, as the robot's confidence-aware role estimate evolves, the robot automatically shifts between unsafe sets of various sizes. When the observed human motion can be well-described by the general-sum game, the marginalized set of human control profiles will place larger probability mass on those trajectories which are optimal under that model (e.g., if the human is confidently estimated to be a follower, then trajectories where the human slows down to let the robot pass). This in turn \textit{shrinks} the estimated human control bounds used during the BRT computation since the likely maximum and minimum controls are structured according to the general-sum game and not according to what is dynamically feasible for the human to execute. 

However, as the human's behavior deviates from the modelled structure, the robot's model confidence will be low for \textit{all} human roles; as all the probability mass concentrates on low values of $\beta$, the marginalized distribution from \eqref{eq:htraj_marginal_dist} approaches a uniform distribution over all trajectories (irrespective of the leader-follower structure). This automatically \textit{enlarges} the human's control bounds (and the resulting BRT) since the distribution indicates that the likely control trajectories could exhibit any dynamically-feasible behavior. 


\section{Experimental Setup}
\label{sec:experimental_setup}


\noindent \textbf{Human-robot system dynamics.}
In our simulation experiments, we consider a dynamical system that encodes the relative dynamics between the robot car and the human-driven car in a pairwise interaction \cite{leung2020infusing, li2020prediction}. The human-driven car is modelled as an extended unicycle model and the robot car is modelled with a high-fidelity bicycle model \cite{rajamani2011vehicle}. The state of the relative system is $\xrel = \big[p^x_\rel, p^y_\rel, \psi_\rel, v_\mathrm{R}, v_\mathrm{H}\big]^\top$, where $p^x_\rel, p^y_\rel$ are the $x$ and $y$-coordinate of the human-driven car in the coordinate frame centered at the geometric center of the robot car with $x$-axis aligned with the heading of the self-driving car, $\psi_\rel$ denotes the relative heading between the two cars, and $v_\mathrm{R}$ (resp., $v_\mathrm{H}$) denotes the speed of the robot car (resp., human-driven car). 
Let $\uR = \big[a_\mathrm{R}, \delta_f\big]$ be the robot's control input where $a_\mathrm{R}$ is the acceleration and $\delta_f$ is the front wheel rotation and $\uH = \big[a_\mathrm{H}, \omega_\mathrm{H}\big]$ denote the human's control input where $a_\mathrm{H}$ is the acceleration and $\omega_\mathrm{H}$ is angular speed. 
The evolution of the relative system is governed by the differential equation $\dot{x}_\rel = \reldyn\big(\xrel, \uR, \uH\big)$ where:
$\dot{p}^x_\rel = \frac{v_\mathrm{R} p^y_\rel}{l_r}\sin(\beta_r) + v_\mathrm{H}\cos(\psi_\rel) - v_\mathrm{R}\cos(\beta_r)$, 
$\dot{p}^y_\rel = -\frac{v_\mathrm{R} p^x_\rel}{l_r}\sin(\beta_r) + v_\mathrm{H}\sin(\psi_\rel) - v_\mathrm{R}\sin(\beta_r)$, 
$\dot{\psi}_\rel = \omega_\mathrm{H} - \frac{v_\mathrm{R}}{l_r}\sin(\beta_r)$,
$\dot{v}_\mathrm{R} = a_\mathrm{R}$, and
$\dot{v}_\mathrm{H} = a_\mathrm{H}$.
Here, $l_f$ (resp., $l_r$) denotes the front (resp., rear) axle length of the robot car and $\beta_r$ is computed via $\beta_r = \tan^{-1}(\frac{l_r}{l_r+l_f}\tan(\delta_f))$.
    
\smallskip
\noindent \textbf{Human reward function.} We model the human's reward function as a linear combination of predefined features \cite{ng2000algorithms}, including the human's: 
1) \textit{speed}: desire to reach the speed limit; 
2) \textit{comfort}: preference for smooth motions; 
3) \textit{reference path deviation}: tendency to follow a reference path (e.g., center lane) in structured roads;
4) \textit{progress}: desire to reach their goal state;
5) \textit{safety}: collision-avoidance objective. 
The weights on these features are obtained by inverse reinforcement learning \cite{ziebart2008maximum} on a human diving data set \cite{interactiondataset}. More details can be found in \cite{sun2021complementing}.


\smallskip
\noindent \textbf{Hyper-parameters.} The horizon used when solving the HJI-VI from \eqref{eq:hji_vi} and the general-sum games \eqref{eq:olgame_hlead_rfollow} and \eqref{eq:olgame_hfollower_rlead} is $\thor = 2 [s]$. We discretize the model confidence parameter into $\beta \in \{0.1, 1, 20\}$, and choose the threshold for rejecting unlikely human control trajectories to be $\epsilon = 0.001$. 

\section{Simulated Human-Robot Interaction Results}
\label{sec:sim_human_robot_results}

We first investigate two core aspects of our proposed method and the overall human-robot system: (a) the ability of our Bayesian BRT to modulate its size based on the confidence-aware game-theoretic model given a range of simulated human behavior, (b) the effect of our safety method on overall robot behavior. We additionally perform two ablation studies to assess the value of incorporating both model confidence and influence models into safety monitors.

\smallskip
\noindent \textbf{Robot planner.} The robot plans via model-predictive control \cite{camacho2013model} and leverages a game-theoretic predictive model of the human behavior as in \cite{sadigh2016planning, fisac2019hierarchical, schwarting2019social}.
At each time step, the robot computes an open-loop control trajectory $\uRtrajopt$ by solving: $\max_{\uRtraj} \mathbb{E}_{\uHtraj} \big[ R_{\mathrm{R}}(\jointstate^0, \uRtraj, \uHtraj) \mid P(\uHtraj \mid \jointstate^0, \uRtraj) \big]$ where the conditional expectation is taken over game-theoretic human trajectories following \eqref{eq:olgame_hfollower_rlead} with a fixed rationality. The robot applies the first control in the trajectory and re-plans at the next time in a receding-horizon fashion. Note that as per prior work in game-theoretic planning, the human is assumed to have a fixed follower role in the interaction at all times. While this introduces some modelling error, it nevertheless enables interesting robot behaviors and is an example of how not all motion planners will be perfect, underscoring the need for monitoring safety of the actual human-robot system.


\smallskip
\noindent \textbf{Safety controller.}
Given a BRT, the robot employs a switching control strategy to avoid unsafe situations \cite{bansal2017hamilton}; this strategy is agnostic to the upstream robot planner which may be arbitrarily complex. Specifically, when the human-driven car reaches the boundary of the BRT, the planner controls are overridden by the safety controller from \eqref{eq:hj_opt_control}. 

\smallskip
\noindent \textbf{Simulated humans.} We simulate three types of humans: (\textit{modeled}) a rational Stackelberg human, (\textit{noisy}) a suboptimal Stackelberg human, and (\textit{unmodeled}) a non-Stackelberg human driving with constant controls (e.g., distracted driver).

\smallskip
\noindent \textbf{Metrics.} We evaluate the safety performance, conservatism, and overall reward of the robot's motion plan when relying on various safety monitors. We measure: 1) \textit{collision rate (CR)}: average number interactions in which the human and robot collide; 2) \textit{safety override rate (SOR)}: average number of time steps during a finite-horizon trajectory when the safety controller is activated; 3) \textit{reward improvement percent (RIP)}: the percent reward increase in the robot's executed trajectory when it uses our Bayesian BRT as compared to using a baseline BRT method (where the appropriate baseline depends on our case study). Mathematically, for any baseline safety method $i$, $\mathrm{RIP}(i) := \frac{R_\mathrm{R}(\jointstate^0, \uRtraj^\mathrm{Bayes}, \uHtraj^\mathrm{Bayes}) - R_\mathrm{R}(\jointstate^0, \uRtraj^i, \uHtraj^i)}{|R_\mathrm{R}(\jointstate^0, \uRtraj^i, \uHtraj^i)|},$ where $x^0$ is the initial condition of the simulation, $\uRtraj^\mathrm{Bayes}$ is the robot's executed control when using our Bayesian BRT, $\uRtraj^i$ is the robot's executed trajectory when using the baseline BRT, and $\uHtraj^i, \uHtraj^\mathrm{Bayes}$ denote the corresponding human trajectories. 

{\renewcommand{\arraystretch}{1}
\begin{table}[t!]
\centering
%
\begin{subtable}{\columnwidth}
\resizebox{\columnwidth}{!}{%
\begin{tabular}{|l|l|l|l|l|l|}
\hline
    \rowcolor{medium_gray}
    \multicolumn{6}{|c|}{\textbf{Round-about Merging Scenario}} \\ \hline \rowcolor{light_gray} 
    & \multicolumn{2}{|c|}{Full BRT} & \multicolumn{2}{|c|}{Bayes BRT} & \\ \hline \rowcolor{light_gray} 
    Human type        & CR           & SOR    & CR  &  SOR        & RIP(Full)     \\ \hline
    \textit{modeled} & 0 & 23.3 & 0 & 4.7 & \textbf{27.75 $\pm$ 4.03}    \\ \hline
    \textit{noisy} & 0 & 29.8 & 0 & 7.3 & \textbf{18.26 $\pm$ 3.96}     \\ \hline
    \textit{unmodeled} & 0 & 42.1 & 0 & 41.7 & 0.06 $\pm$ 0.19    \\ \hline
\end{tabular}}
\end{subtable}

\vspace{2mm}
%
\begin{subtable}{\columnwidth}
\resizebox{\columnwidth}{!}{%
\begin{tabular}{|l|l|l|l|l|l|}
\hline
    \rowcolor{medium_gray}
    \multicolumn{6}{|c|}{\textbf{Highway Scenario}} \\ \hline \rowcolor{light_gray} 
    & \multicolumn{2}{|c|}{Full BRT} & \multicolumn{2}{|c|}{Bayes BRT} & \\ \hline \rowcolor{light_gray} 
    Human type        & CR           & SOR    & CR  &  SOR        & RIP(Full)     \\ \hline
    \textit{modeled} & 0 & 28.3 & 0 & 9.2 & \textbf{24.26 $\pm$ 6.16}     \\ \hline
    \textit{noisy} & 0 & 43.2 & 0 & 17.4 & \textbf{14.83 $\pm$ 4.22}     \\ \hline
    \textit{unmodeled} & 0 & 64.8 & 0 & 62.3 & 0.13 $\pm$ 0.08   \\ \hline
\end{tabular}}
\end{subtable}
\caption{Results for the round-about and highway; each row is averaged across 20 interactions. CR: collision rate; SOR: safety override rate; RIP: reward improvement \%.} 
\label{tab:bayes_vs_full_simulated_results}
\end{table}}

\begin{figure}[th!]
\begin{center}
\begin{picture}(300, 80)
\put(30,  0){\epsfig{file=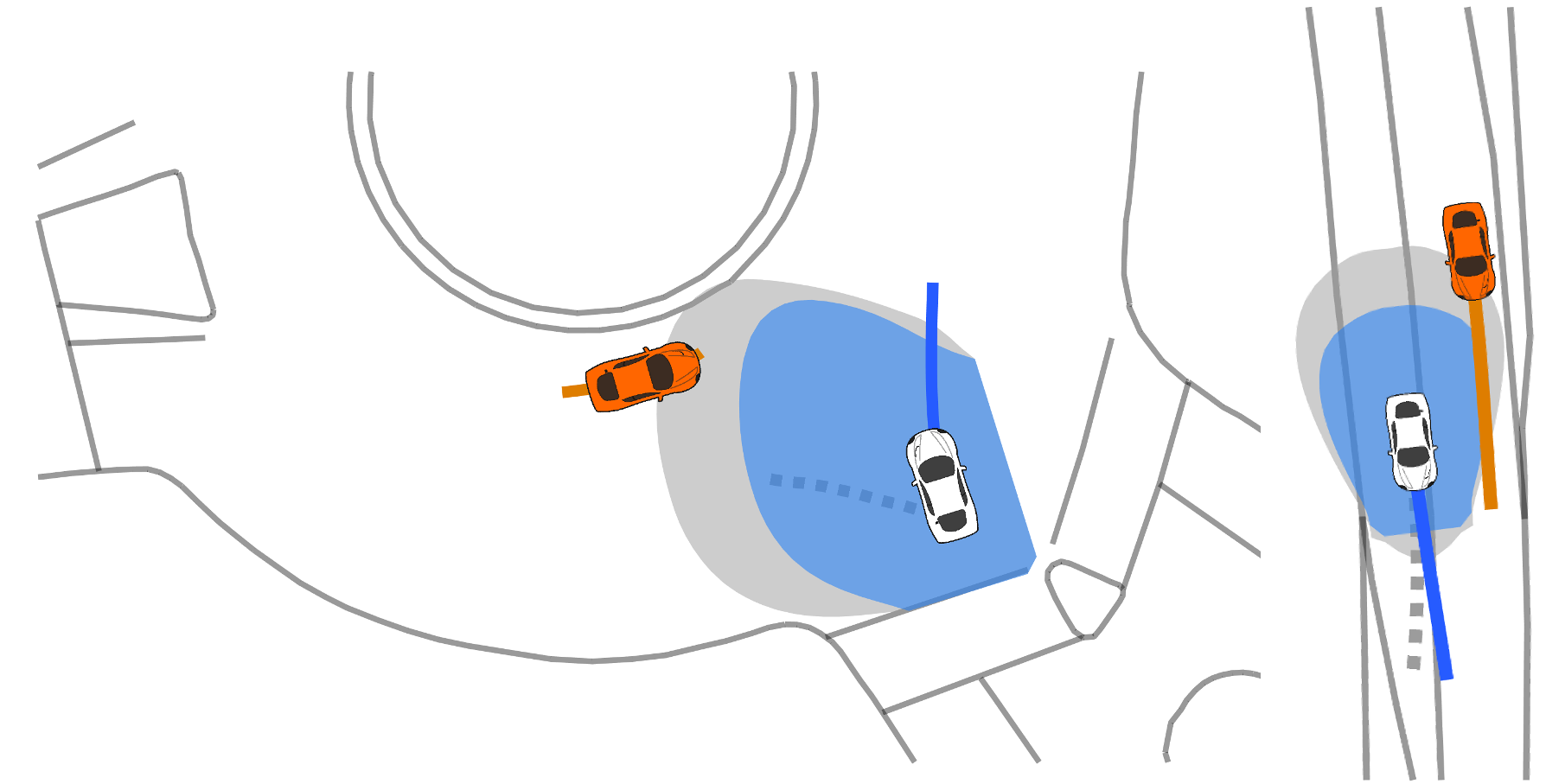, width = 0.8\linewidth, trim=4cm 4cm 0.5cm 4cm,clip}}  
\put(20,80){(a)}
\put(180,80){(b)}
\end{picture}
\end{center}
\vspace{-0.4cm}
\caption{Robot car interacts with a simulated human \textit{follower}. Our Bayes BRT infers that the human is game-theoretic and tends to cooperate; it reduces the size of the unsafe set (filled blue) accordingly. A 2-D slice of the full 5-D unsafe set is visualized in the $p^x_{rel}$ and $p^y_{rel}$ dimensions with all other states held fixed at the current values. Dotted gray line is the robot's trajectory when using the full BRT (filled gray). Note that the full BRT results in the robot executing unnecessary safety maneuver in (a) and aborting merging in (b).}
\label{fig:round_about_and_highway}
\vspace{-0.6cm}
\end{figure}

\subsection{On the utility of the Bayesian BRT vs. full BRT}
\label{sec:sim_h_results_full_vs_bayes_brt}

We first compare the performance of our Bayesian BRT to the full BRT (where the unsafe sets are computed with respect to all dynamically-feasible human controls) for each type of simulated human. 
We compute our metrics in two traffic scenarios, round-about merging and highway lane-change, across 20 interactions for each test. 
Our findings for both scenarios are summarized in Tab.~\ref{tab:bayes_vs_full_simulated_results} and snapshots of the full BRT and our Bayesian BRT are visualized in Fig.~\ref{fig:round_about_and_highway}.

Results from both scenarios indicate that when the true simulated human behaves according to the general-sum Stackelberg game, our Bayesian BRT can detect this modelled structure and reduce the conservatism of the safety guarantee (e.g., in the round-about, resulting in $\sim$20\% less safety controller activations than the full BRT) while preserving collision-free human-robot interaction. 
Furthermore, the overall reward of the robot's executed trajectory is increased by 27.75\% when compared to the trajectory the robot would execute if it was relying on the overly-conservative full BRT as its safety monitor. Importantly, as the simulated human behavior becomes increasingly misspecified, we also see the Bayesian BRT increasing in conservatism---ultimately approaching comparable performance across all metrics to the full BRT when the distracted human driver (\textit{unmodeled}) behaves in a completely non-game-theoretic way.

\subsection{Ablation Study 1: On the value of model confidence}
\label{sec:sim_h_results_model_confidence}

\begin{figure}[th!]
\begin{center}
\begin{picture}(300, 60)

\put(0,  0){\epsfig{file=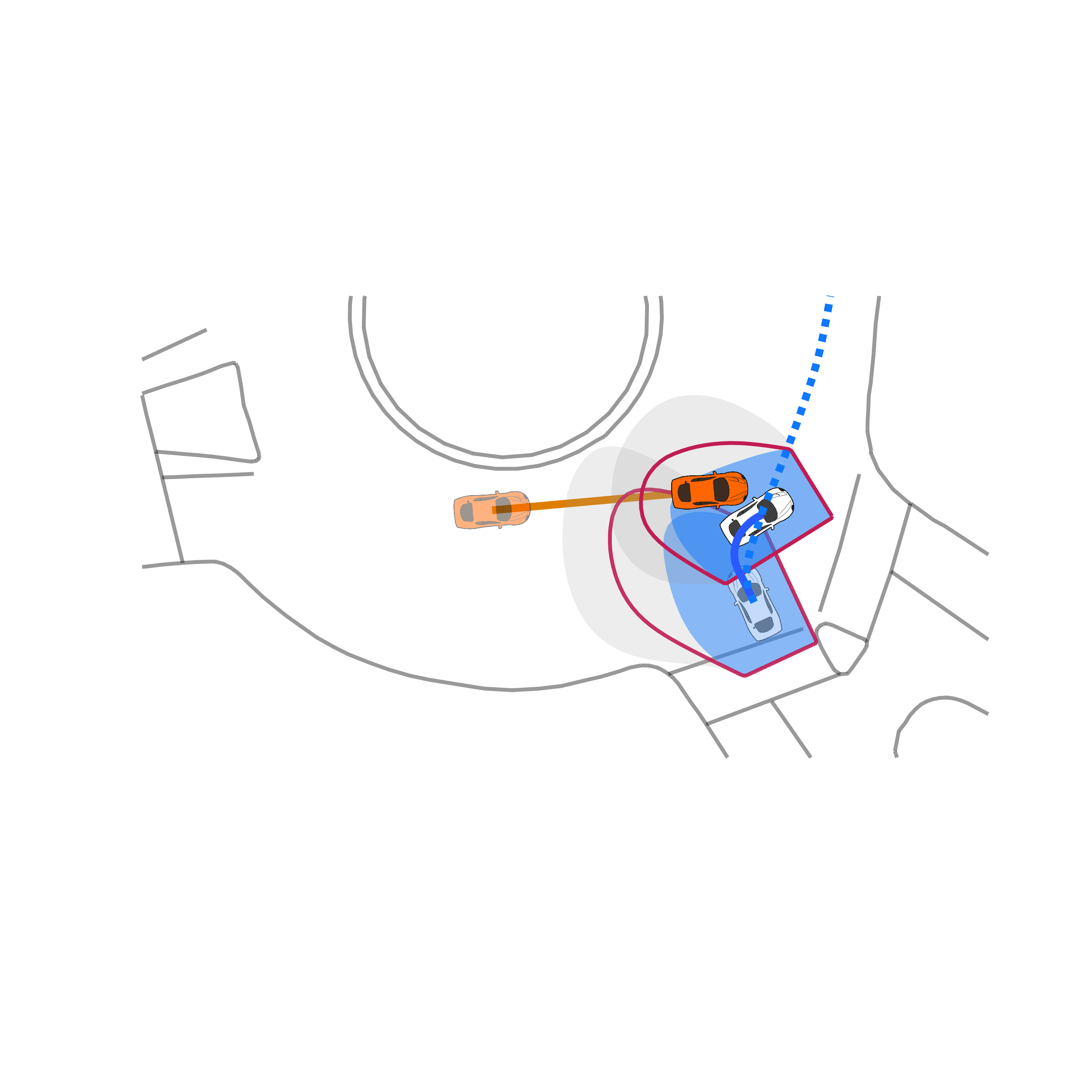, width = 0.5\linewidth, trim=4cm 6cm 0.5cm 5cm,clip}}  
\put(130,  0){\epsfig{file=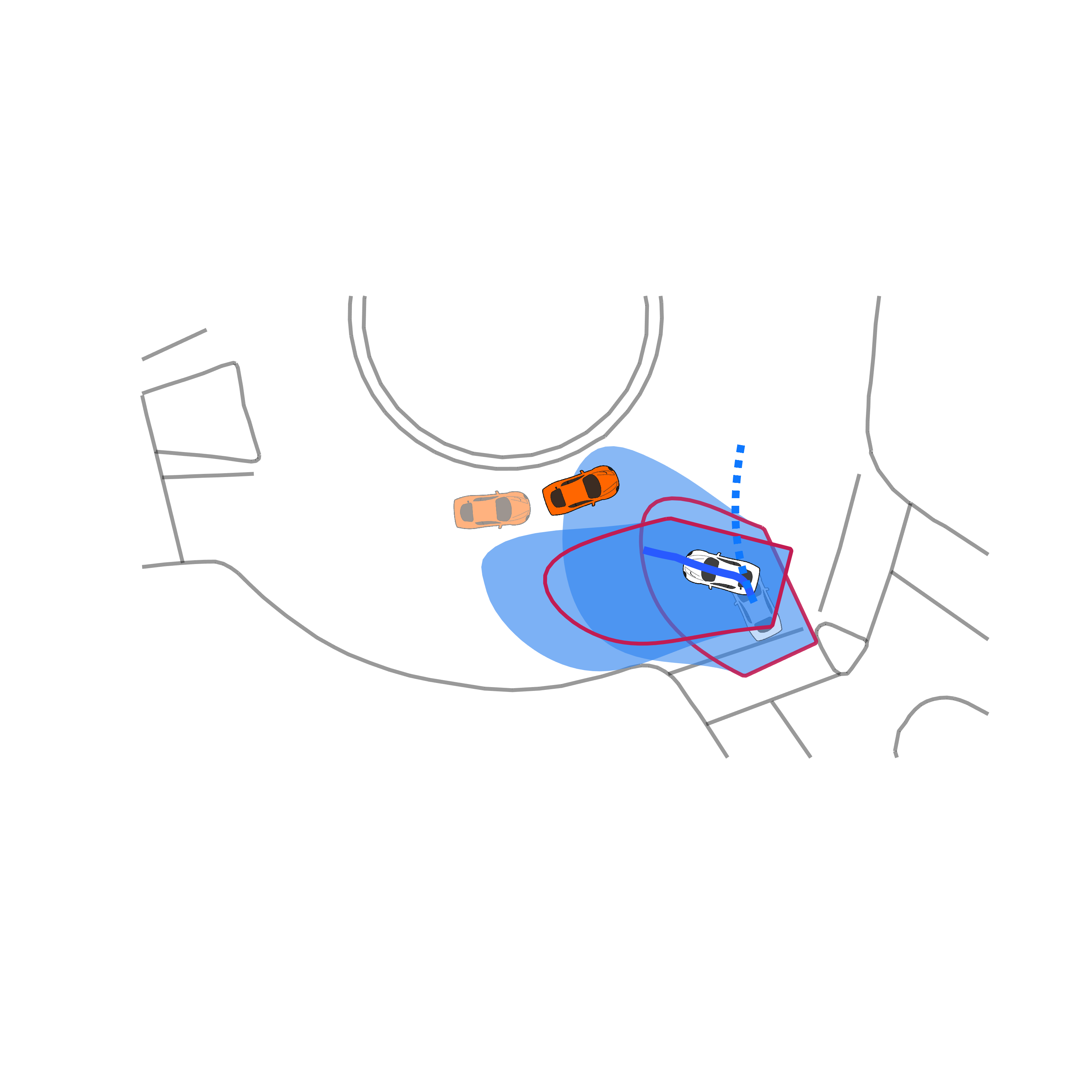, width = 0.5\linewidth, trim=4cm 6cm 0.5cm 5cm,clip}}  

\put(0,60){(a)}
\put(130,60){(b)}

\end{picture}
\end{center}
\caption{(a) Failure to detect model mismatch results in collision. The $\lambda$-only BRT (filled blue) is smaller than the least-conservative BRT (pink outline, defined in Section \ref{sec:traffic_data_results}), meaning safety is no longer guaranteed. Dotted blue line is robot trajectory using our Bayes BRT, avoiding collision. (b) Failure to use an interactive human model results in the robot avoiding the human even when they try to cooperate.}
\label{fig:case_studies}
\vspace{-0.6cm}
\end{figure}

To investigate the effect of model confidence in our method,
we implement a version of our Bayesian BRT without model confidence but with the general-sum model; we call this method \textit{$\lambda$-only Bayesian BRT}. We stress-test this model's utility in the round-about scenario when the robot interacts with a non-game-theoretic distracted human (\textit{unmodeled}). As Tab.~\ref{tab:lambda_only_brt} indicates, trusting the model completely leads to a high collision percentage; this is because the estimated control bounds do not adequately capture the true human controls (see visualization in left of Fig.~\ref{fig:case_studies}). 
Additionally, analyzing the reward improvement percent that our Bayesian BRT achieves when compared to the $\lambda$-only Bayesian BRT, we see a 9.18\% improvement since the robot detects misspecification early and takes evasive maneuvers.

{\renewcommand{\arraystretch}{0.8}
\centering
\begin{table}[h!]
\resizebox{\columnwidth}{!}{%
\begin{tabular}{|l|l|l|l|}
    \hline
    \rowcolor{light_gray}
    & \multicolumn{2}{|c|}{$\lambda$-only Bayesian BRT} & \\ \hline \rowcolor{light_gray} 
    Human type        & CR           & SOR        & RIP($\lambda$-only)     \\ \hline
    \textit{unmodeled} & 25 & 11.3 &  \textbf{ 9.18$\pm$2.9}   \\ \hline
\end{tabular}}
\caption{Results without $\beta$ but with game-theoretic model.} 
\label{tab:lambda_only_brt}
\end{table}
}

\subsection{Ablation Study 2: On the value of game-theoretic models}
\label{sec:sim_h_results_stackelberg}

To understand the utility of general-sum game-theoretic models in our safety monitor, we implement a version of our Bayesian BRT with a human-in-isolation model but with model confidence (as in \cite{fisac2018probabilistically}): we call this method \textit{$\beta$-no game Bayesian BRT}.
We stress-test this method against a simulated human which behaves rationally according to the game-theoretic interaction model. Results are shown in Tab.~\ref{tab:beta_only_brt}.
Because the underlying human model in \textit{$\beta$-no game Bayesian BRT} assumes that people are not influenced by the robot, then the model confidence is always low in highly dynamic interactions with the human (see right of Fig.~\ref{fig:case_studies}). This results in the safety monitor relying on the full set of human controls and, although the robot never collides, it activates its safety monitor near identically as often as the full BRT interacting with a \textit{modeled} human from Tab.~\ref{tab:bayes_vs_full_simulated_results}. 
Our Bayesian BRT improves the robot's reward by $\sim$29\% over the \textit{$\beta$-no game Bayesian BRT} since the robot can confidently execute its plan without the safety monitor intervening. 

{\renewcommand{\arraystretch}{1}
\centering
\begin{table}[h!]
\resizebox{\columnwidth}{!}{%
\begin{tabular}{|l|l|l|l|}
\hline
    \rowcolor{light_gray}
    & \multicolumn{2}{|c|}{$\beta$-no game Bayes BRT} & \\ \hline \rowcolor{light_gray} 
    Human type        & CR           & SOR        & RIP($\beta$-no game)     \\ \hline
    \textit{modeled} & 0 & 24.7 & \textbf{29.18 $\pm$ 3.63}      \\ \hline
\end{tabular}}
\caption{Results with $\beta$ but no game-theoretic model.} 
\label{tab:beta_only_brt}
\vspace{-0.4cm}
\end{table}
}

\section{Evaluation with Real Traffic Data}
\label{sec:traffic_data_results}


We investigate how the full BRT and our Bayesian BRT perform \textit{ex post facto} on recorded human traffic data.
We extract $200$ pair-wise interactions from \cite{interactiondataset} and for each interaction we assign one car as the robot car (the other as the human), and run our approach of constructing the Bayesian BRT while replaying the two cars' recorded trajectories. 
We compare our method to two baselines: (1) the full BRT and (2) the least-conservative BRT, $\truebrs(\tau)$, which is obtained by restricting the human's control set in the HJI-VI to be the maximum and minimum controls observed in a $T$-length snippet of the ground-truth human trajectory starting at the current timestep. Theoretically, this is the least conservative unsafe set that can be obtained via our approach.  

\begin{figure}[t!]
\begin{center}
\begin{picture}(300, 60)

\put(-10, 0){\epsfig{file=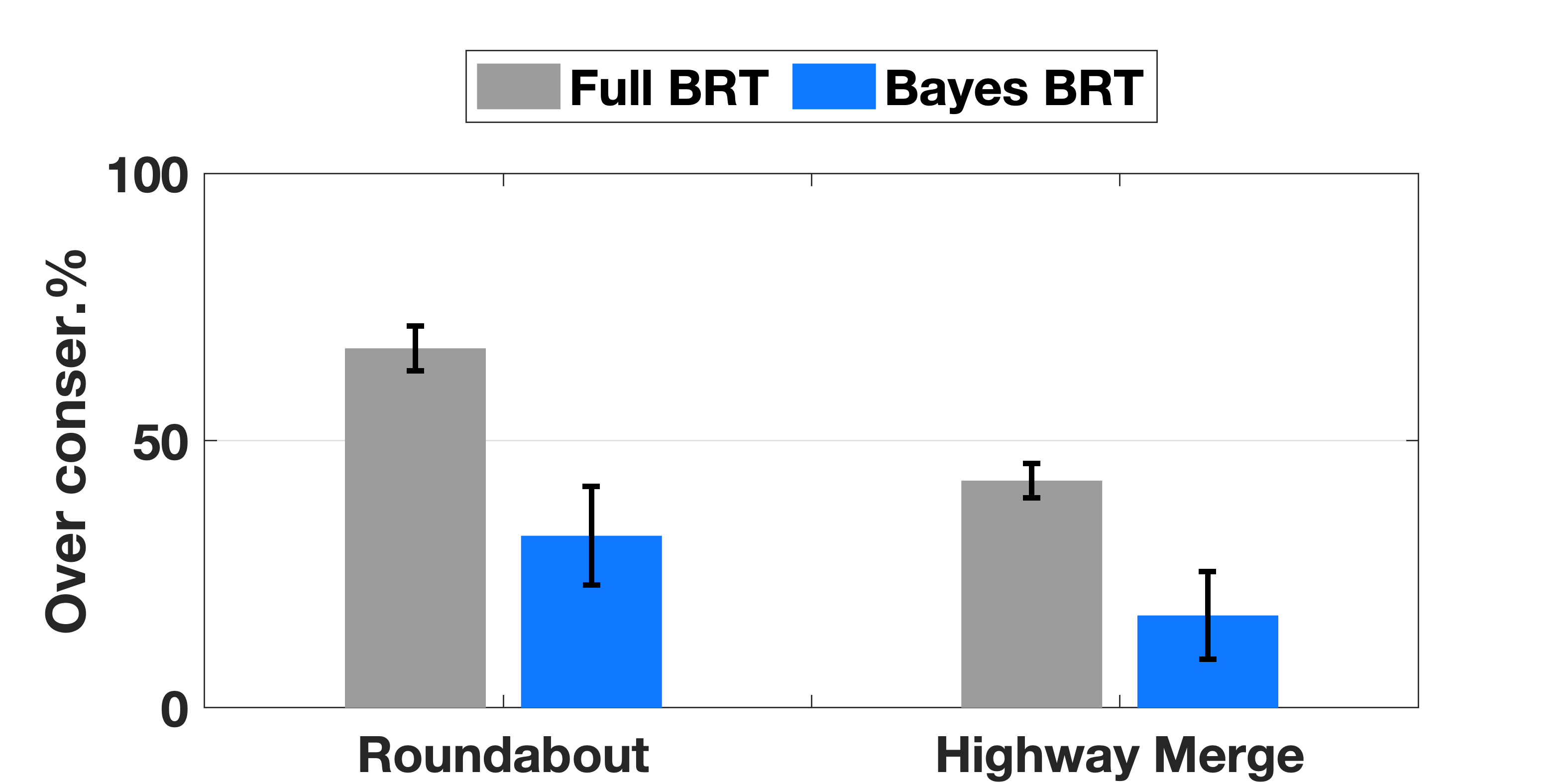, width = 0.55\linewidth, trim=0cm 0cm 1cm 0.5cm,clip}}
\put(120, 0){\epsfig{file=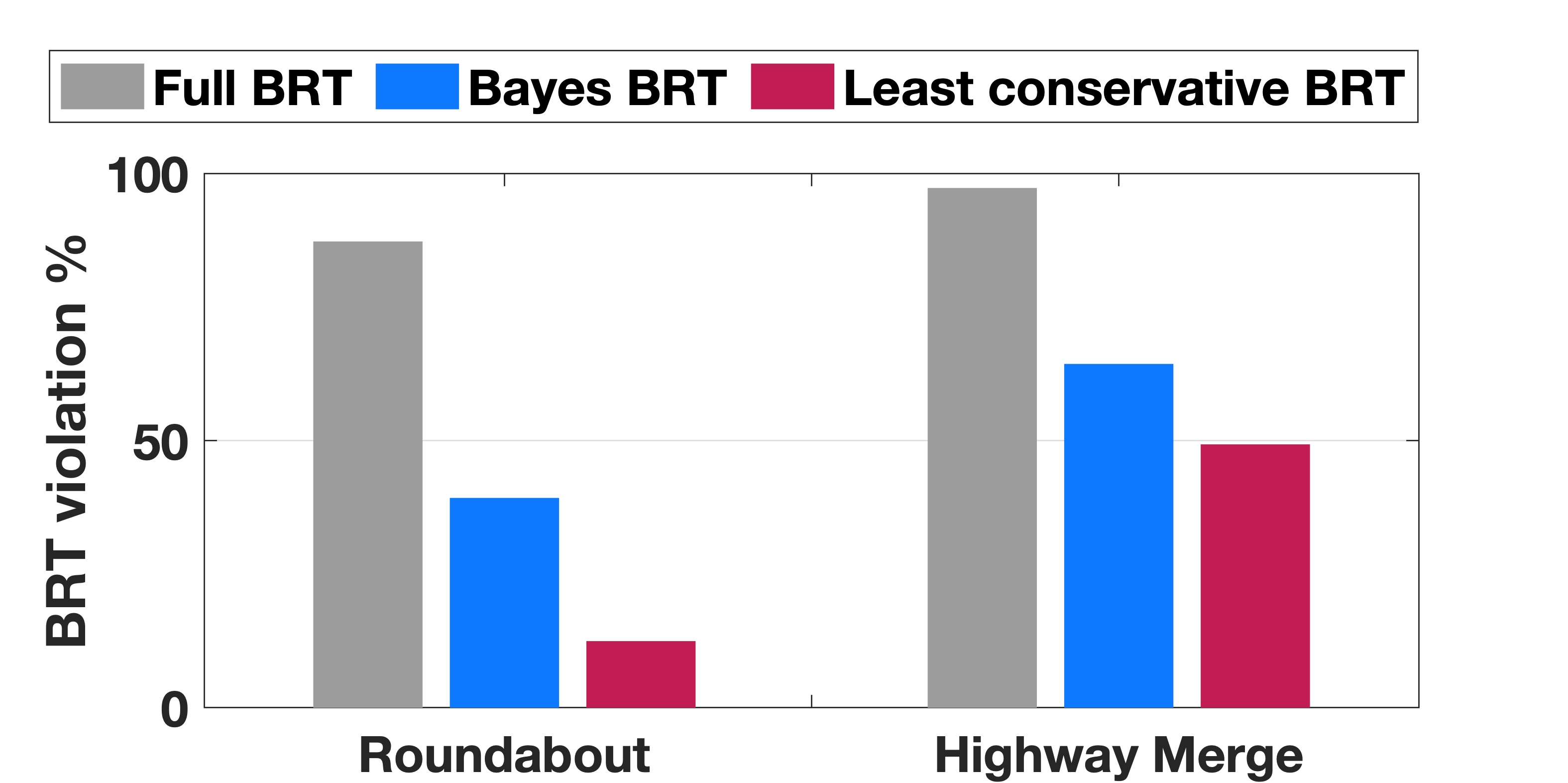, width = 0.55\linewidth, trim=0cm 0cm 1cm 0.5cm,clip}}  


\end{picture}
\end{center}
\caption{(left) Our method is not overly-conservative compared to the least-restrictive BRT. (right) Our method decreases safety violations vs. the full BRT.}
\label{fig:BRT_eval}
\vspace{-2em}
\end{figure}

At each timestep, we measure the \textit{over-conservatism percentage} of a given BRT $\brs^i(\tau), i \in \{\mathrm{Full}, \mathrm{Bayes}\}$, by comparing the area of $\brs^i(\tau)$ to the least-conservative BRT $\truebrs(\tau)$. Mathematically, this is $\frac{\mathcal{A}(\brs^i(\tau)) - \mathcal{A}(\truebrs(\tau))}{\mathcal{A}(\truebrs(\tau))}$, where the function $\mathcal{A}$ maps a set of states to its geometric area. We also measure the \textit{BRT violation percentage}: the percentage of interactions flagged as unsafe by a given BRT method. Note that once the BRT is breached, trajectory replay stops.

Our findings are summarized in \cref{fig:BRT_eval}, showing the average over-conservatism percentage and average BRT violation percentage for each traffic scenario. Both the over-conservatism and the BRT violation percentage of our Bayesian BRT are significantly reduced compared to the full BRT. This suggests that human driving is closer, from a safety perspective, to our method than to the traditionally used full BRT approach. However, humans are still violating the constraints our method would impose, meaning there is still room to improve efficiency while preserving the same safety levels as humans.
Finally, note that the over-conservatism percentage of our Bayesian BRT is positive, indicating that our Bayesian BRT is more conservative than the least-conservative BRT and thus preserves the quality of our safety monitor.

\section{Discussion \& Conclusion}
\label{sec:discussion}

We proposed that robot safety monitors be imbued with confidence-aware game-theoretic models. By restricting the set of feasible human controls based on how much the human follows the game-theoretic model, the robot can automatically interpolate between smaller unsafe sets consistent with the human model and the full worst-case unsafe set.

Our traffic data experiments revealed that even the least-conservative BRT is more conservative than real drivers; this is due to our control bound construction and the zero-sum nature of the reachability game. We are excited for future work on new safety methods which reflect human notions of safety. 
Further, as in \cite{leung2020infusing}, designing robot planners which are aware of the safety monitor is an interesting future direction.


\balance
\bibliographystyle{IEEEtran}
\bibliography{references}
\end{document}